%% file: main.tex
\documentclass[11pt]{article}
\PassOptionsToPackage{numbers, compress}{natbib}
\usepackage[preprint]{neurips_2026}
\usepackage[utf8]{inputenc}
\usepackage[T1]{fontenc}
\usepackage{amsmath,amssymb,amsfonts}
\usepackage{graphicx}
\usepackage{booktabs}
\usepackage{multirow}
\usepackage{xcolor}
\usepackage{hyperref}
\usepackage{enumitem}
\usepackage{subcaption}
\usepackage{tabularx}
\usepackage{array}
\usepackage{fancyvrb}
\usepackage{wrapfig}

\hypersetup{colorlinks=true,linkcolor=blue,citecolor=blue,urlcolor=blue}

\newcommand{\benchmark}{MMCL-Bench}

\title{\benchmark{}: Multimodal Context Learning from Visual Rules, Procedures, and Evidence}

\author{
Yifan Chen$^{1,*}$ 
\hspace{6mm} 
Fei Yin$^{1,*}$ \hspace{6mm} Qingyan Bai$^{2}$ \hspace{6mm} Zicheng Lin$^{3,\dagger}$ \hspace{6mm}  Yujiu Yang$^{3,\dagger}$ \\
$^{1}$University of Cambridge\hspace{3mm}  $^{2}$HKUST \hspace{3mm} $^{3}$Tsinghua University 
\\
}

\date{}

\begin{document}

\maketitle

\let\thefootnote\relax\footnotetext{$^*$ Equal contribution.} 
\footnotetext{$^\dagger$  Corresponding authors.}

\begin{center}
\begin{minipage}{0.95\linewidth}
    \centering
    \includegraphics[width=\linewidth]{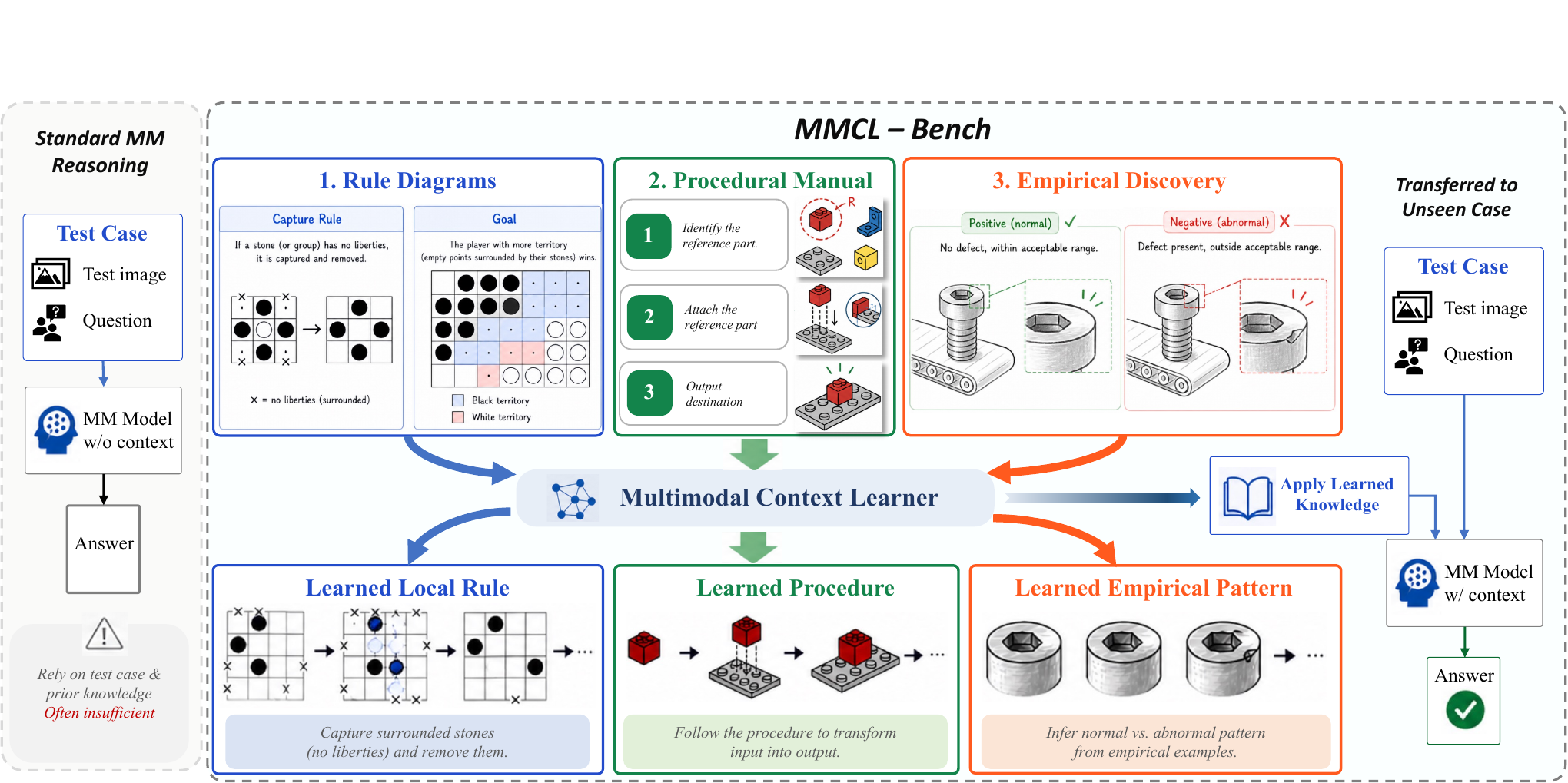}
    \captionof{figure}{
    Standard multimodal reasoning often answers from the  prior knowledge, whereas \benchmark{} requires learning  from multimodal teaching context and transferring them to an unseen visual test instance. 
    Success depends on context learning rather than on direct image-question answering.
    }
    \label{fig:teaser}
\end{minipage}
\end{center}

\input{sections/0_abstract}

\input{sections/1_intro}
\input{sections/2_related_work}
\input{sections/3_benchmark}
\input{sections/4_results}
\input{sections/5_analysis}
\input{sections/7_conclusion}

\bibliographystyle{unsrtnat}
\bibliography{references}

\input{sections/8_appendix}

\end{document}

%% file: sections/0_abstract.tex
\begin{abstract}
We introduce \benchmark{}, a benchmark for multimodal context learning: learning task-local rules, procedures, and empirical patterns from visual or mixed-modality teaching context and applying them to new visual instances. Unlike text-only context learning or standard multimodal question answering, this setting requires models to recover and localize relevant evidence from images, screenshots, manuals, videos, and frame sequences before they can reason over the learned context. \benchmark{} contains 102 tasks spanning three categories: rule system application, procedural task execution, and empirical discovery and induction. We evaluate frontier multimodal models with strict rubric-based scoring and find that current systems remain far from robust multimodal context learning, with even the strongest model solving fewer than one-third of tasks under strict evaluation. Diagnostic ablations and error analysis show that failures arise throughout the context-to-answer pipeline, including 
context anchoring, 
visual evidence extraction, context reasoning, and response construction. \benchmark{} thus highlights multimodal context learning as an important unsolved capability bottleneck for current multimodal models.
\end{abstract}


%% file: sections/1_intro.tex
\section{Introduction}
\label{sec:intro}

Context learning is the ability to acquire task-specific knowledge from provided context and apply it to a new problem. Text-only benchmarks such as CL-Bench~\cite{dou2026clbench} evaluate whether language models can read a local rulebook, manual, policy, or documentation snippet and then solve tasks that depend on that context \citep{dou2026clbench,clbenchwebsite2026}. Many real context-learning problems, however, are not text-only. A model may need to assemble furniture from illustrated manuals, interpret professional rules from diagrams and examples, detect anomalies in industrial inspection images or videos.
Existing text-only benchmarks therefore miss key challenges of multimodal context learning.

We study \emph{multimodal context learning} (MMCL): learning a task-local rule, procedure, or empirical pattern from visual or mixed-modality teaching context and transferring it to a new visual test instance. As illustrated in Fig.~\ref{fig:teaser}, this setting differs from standard multimodal question answering because the operative knowledge is not assumed to be in the model's parametric memory or in the test image alone. It is contained in the provided context. To succeed, a model must read the teaching evidence, localize the relevant part of that evidence, infer the local rule or procedure, bind it to the test state, and produce an answer satisfying fine-grained requirements.

This shift makes perception and grounding part of context learning itself. In text-only settings, the context is already available as tokens. In MMCL, the model must first convert diagrams, screenshots, manuals, videos, frame sequences, or visual examples into a usable intermediate state. A single mistaken color, coordinate, object identity, manual step, or state update can produce an answer that is coherent under the wrong internal state but incorrect under the task rubric. Thus, failures are often not simply cases where the model ignores context; they are cases where the model uses a misread or mislocalized version of the context.

We introduce \benchmark{}, a benchmark designed to measure this capability. The benchmark contains 102 tasks across three categories: Rule System Application, Procedural Task Execution, and Empirical Discovery and Induction. Tasks span both controlled rule-learning settings and practical scenarios, including visual formalisms, procedural manuals, and real-world induction tasks.

We construct tasks through a taxonomy-guided human--agent workflow. Candidate tasks are generated or curated so that the teaching context contains the task-local source of truth and the test context requires transfer to a new case. We filter tasks that can be solved without the teaching context or that rely too heavily on familiar priors. Evaluation uses strict all-rubric task scoring together with rubric-level accuracy. To avoid introducing additional multimodal perception errors during judging, the judge receives only the model response and the task rubrics, not the original multimodal context.

The resulting benchmark exposes a large gap between current multimodal systems and robust MMCL, even under strong API settings with extended reasoning. The central weakness is not only final rule application, but the inability to reliably recover and anchor the task-local source of truth from multimodal teaching context. Failures are rarely all-or-nothing: models often satisfy many local rubrics but miss one crucial coordinate, manual step, state update, or rule condition. This makes \benchmark{} useful not only as a benchmark of end-task success, but also as a diagnostic instrument for studying where multimodal context learning breaks down.

Our diagnostic ablations further show that failures span four stages of the MMCL pipeline---visual evidence extraction, context anchoring, context reasoning, and response construction---with the largest bottlenecks arising in the first two. Many errors occur because the model does not extract the right visual state, and another major share occurs when the relevant evidence is hard to localize within long or visually dense context, especially in procedural tasks. Even when visual evidence is textualized, some tasks remain difficult because they require applying local rules over multiple state updates. MMCL therefore raises a broader modeling challenge than final-answer reasoning alone: closing the gap will likely require not only stronger reasoning, but also better visual state extraction, long-context grounding, and intermediate-state verification.


Our main contributions are: (1) we introduce multimodal context learning (MMCL) and present \benchmark{}, a 102-task benchmark for studying how models learn task-local rules, procedures, and patterns from multimodal context and transfer them to new instances; (2) we show that current multimodal models struggle with MMCL, and provide diagnostic analyses that reveal distinct failure modes in perception, grounding, and reasoning.

%% file: sections/2_related_work.tex
\section{Related Work}
\label{sec:related}

\vspace{2pt}
\noindent \textbf{Context learning benchmarks.}
\benchmark{} is  closely related to text-based context learning benchmarks such as CL-Bench~\citep{dou2026clbench,clbenchwebsite2026}, which evaluate whether models can learn task-specific knowledge from provided text context. More broadly, long-context benchmarks study retrieval and integration over long textual or multimodal inputs~\citep{bai2024longbench,li2024longiclbench,wang2025mmlongbench,song2024milebench}, while multimodal ICL benchmarks evaluate adaptation from few-shot demonstrations~\citep{alayrac2022flamingo,zong2024vlicl}. In contrast, \benchmark{} studies context learning from \emph{declarative} multimodal teaching context, where the operative rule, procedure, or empirical pattern must be inferred from visual or mixed-modality materials. This makes perception, grounding, and state tracking part of context learning itself~\citep{tong2024eyes,chen2023shikra}.

\vspace{2pt}
\noindent \textbf{Image-text reasoning benchmarks.}
Image-text benchmarks such as VQA v2 test whether models ground answers in paired images and questions rather than language priors alone~\citep{goyal2017vqav2}. A large follow-up literature studies visually grounded reading and document understanding, including TextVQA, ST-VQA, OCR-VQA, DocVQA, MP-DocVQA, InfographicVQA, and ChartQA~\citep{singh2019textvqa,biten2019stvqa,mishra2019ocrvqa,mathew2021docvqa,tito2022mpdocvqa,mathew2022infographicvqa,masry2022chartqa}. 
Recent benchmarks also probe more specialized multimodal reasoning failures. C-VQA injects counterfactual presuppositions into image questions to test reasoning about altered visual states~\citep{zhang2024cvqa}, MIA-Bench evaluates fine-grained multimodal instruction following~\citep{qian2024miabench}, and MMKC-Bench evaluates cases where visual evidence conflicts with a model's stored knowledge~\citep{jia2025mmkcbench}.
These benchmarks are valuable for measuring grounding, OCR, chart understanding, instruction compliance, and prior-conflict handling, but they are still largely framed as question answering over a single image, chart, or document set. \benchmark{} instead systematically organizes tasks by the required cognitive process, centering practical task-local contexts,
where the model must first infer the local pattern and then apply it to a new case.

\vspace{2pt}
\noindent \textbf{Video and multimodal agent benchmarks.}
Video and agent benchmarks extend multimodal evaluation to longer-horizon and more interactive settings. MT-Video-Bench evaluates multi-turn video dialogue over diverse perceptual and interactive capabilities~\citep{pan2025mtvideobench}. MORSE-500 uses programmatically generated videos to stress-test abstract, physical, spatial, and temporal visual reasoning~\citep{cai2025morse500}. VideoWebArena evaluates multimodal agents that use video tutorials to complete web tasks~\citep{jang2024videowebarena,zhou2023webarena}. Some of these settings implicitly involve context learning, since models must use information from videos, demonstrations, or tutorials. However, they typically do not isolate the setting we study here: learning a task-local rule, procedure, or empirical pattern from multimodal teaching context and transferring it to a new instance. \benchmark{} focuses explicitly on this abstraction-and-transfer bottleneck across both static and dynamic evidence.

\input{tables/0_related_work}

%% file: tables/0_related_work.tex
\begin{table}[t]
    \centering
    \scriptsize
    \caption{Comparison with related benchmarks. ``Task-local ctx.'' denotes context provided within each task instance that is necessary for solving it. ``Grounded setting'' denotes practical task settings such as documents, procedures, or web tasks. ``CL target'' denotes whether learning from provided context is an explicit primary evaluation target rather than an implicit subskill.}

    \label{tab:related_comparison}
    \begin{tabularx}{\linewidth}{l>{\centering\arraybackslash}X>{\centering\arraybackslash}X>{\centering\arraybackslash}X>{\centering\arraybackslash}X>{\centering\arraybackslash}X}
        \toprule
        Benchmark & Image & Video/Seq. & Task-local ctx. & Grounded setting & Context learning target \\
        \midrule
        C-VQA~\citep{zhang2024cvqa} & $\checkmark$ & -- & -- & -- & -- \\
        MMKC-Bench~\citep{jia2025mmkcbench} & $\checkmark$ & -- & -- & -- & -- \\
        MT-Video-Bench~\citep{pan2025mtvideobench} & -- & $\checkmark$ & -- & -- & -- \\
        VideoWebArena~\citep{jang2024videowebarena} & -- & $\checkmark$ & $\checkmark$ & $\checkmark$ & -- \\
        CL-Bench~\citep{dou2026clbench} & -- & -- & $\checkmark$ & -- & $\checkmark$ \\
        \benchmark{} & $\checkmark$ & $\checkmark$ & $\checkmark$ & $\checkmark$ & $\checkmark$ \\
        \bottomrule
    \end{tabularx}
\end{table}

%% file: sections/3_benchmark.tex
\section{Benchmark}
\label{sec:benchmark}

We present \benchmark{}, an evaluation benchmark for measuring multi-modal context learning.
It measures whether models can infer task-specific rules, procedures, and empirical patterns from multimodal teaching context, and then apply them to new visual instances.
Unlike prior benchmarks that either study context learning in text-only settings~\cite{dou2026clbench} or assess multimodal reasoning without requiring local rule induction, \benchmark{} evaluates whether models can learn directly from visual and mixed-modality evidence, including images, screenshots, manuals, videos, and frame sequences. 

\subsection{Task Format}
\label{sec:task_format}
Each task in \benchmark{} includes a teaching context, a test context, a message prompt, and a set of rubrics. The teaching context provides the task-local rule, procedure, or empirical pattern in multimodal form. The test context presents a new instance that requires applying what was learned from the teaching materials. The message prompt specifies the required output without explicitly revealing the underlying rule, and the rubrics define the fine-grained criteria used for evaluation.

The benchmark is designed so that success depends on learning from context rather than recognizing the test input in isolation. To solve a task, the model must recover the operative rule or procedure from the teaching materials and transfer it to a new case.

\subsection{Context Taxonomy}
\label{sec:taxonomy}

\benchmark{} organizes tasks by the form of context learning they require, rather than by surface domain or modality. The benchmark covers 102 tasks across 14 subcategories in three primary categories: Rule System Application, Procedural Task Execution, and Empirical Discovery and Induction.

\begin{wrapfigure}[14]{r}{0.44\columnwidth}
    \vspace{-18pt}
    \centering
    \includegraphics[width=\linewidth]{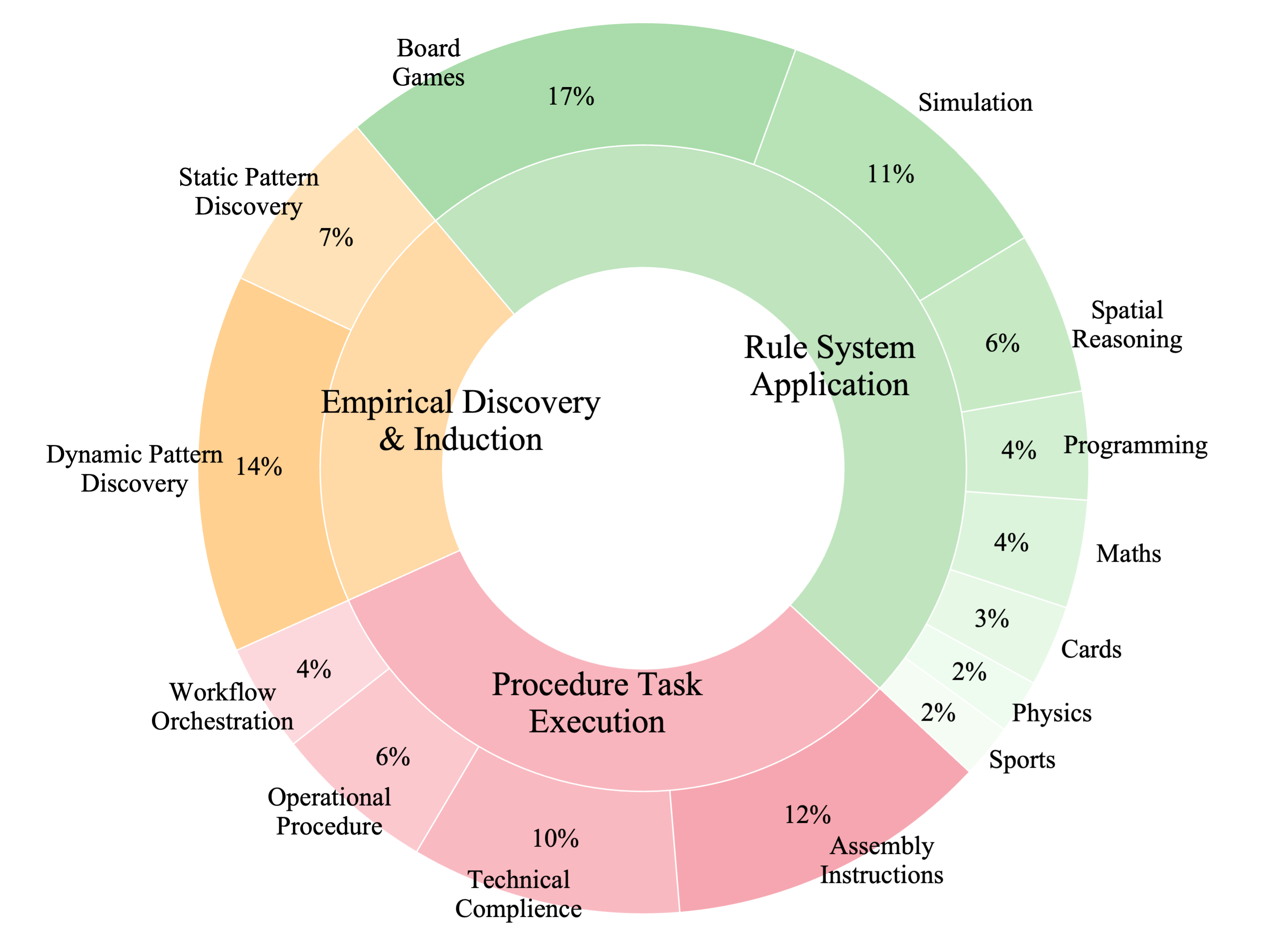}
    \caption{Task distribution across categories.}
    \label{fig:taxonomy_distribution}
    \vspace{-10pt}
\end{wrapfigure}

\vspace{2pt}
\noindent \textbf{Rule System Application.}
The context defines a task-local rule system that must be applied to a new visual state. Unlike text-only rule-learning settings, these rules are often conveyed through diagrams, boards, grids, screenshots, or other structured visual materials, and may conflict with familiar prior knowledge. Models must infer the operative rule from the teaching context and apply it correctly to tasks such as analyzing modified board-game states, reasoning over visual formalisms, interpreting visual DSLs, or solving simulation-like state updates. This category contains eight subcategories: board games, simulation, spatial reasoning, visual DSLs, visual formalisms, cards, physics, and sports.

\vspace{2pt}
\noindent \textbf{Procedural Task Execution.}
The context provides a procedure, workflow, or specification that must be followed to solve a new case. Representative materials include assembly manuals, UI specifications, operational runbooks, and workflow diagrams. Models must identify the relevant step, state, or routing condition from the multimodal context and use it to produce the correct next action, defect list, or decision. This category contains four subcategories: assembly instructions, technical specification compliance, operational procedures, and workflow orchestration.

\vspace{2pt}
\noindent \textbf{Empirical Discovery and Induction.}
In this category, the context does not explicitly state the rule. Instead, models must infer a hidden pattern, feature partition, or state-transition mechanism from multimodal examples and then transfer it to a new instance. The key distinction is the form of the teaching material: static pattern discovery uses fixed visual examples, such as learning how leaf or crystal specimens should be classified from labeled reference images, whereas dynamic pattern discovery uses videos or frame sequences, such as inferring irrigation or animal-growth dynamics from observed state changes over time. Compared with the previous two categories, these tasks place greater weight on inductive inference from multimodal observations rather than direct application of an explicitly given rule or procedure. This category contains two subcategories: static pattern discovery and dynamic pattern discovery.


\subsection{Construction Pipeline}
\label{sec:construction}
\begin{figure}[t]
    \centering
    \includegraphics[width=\linewidth]{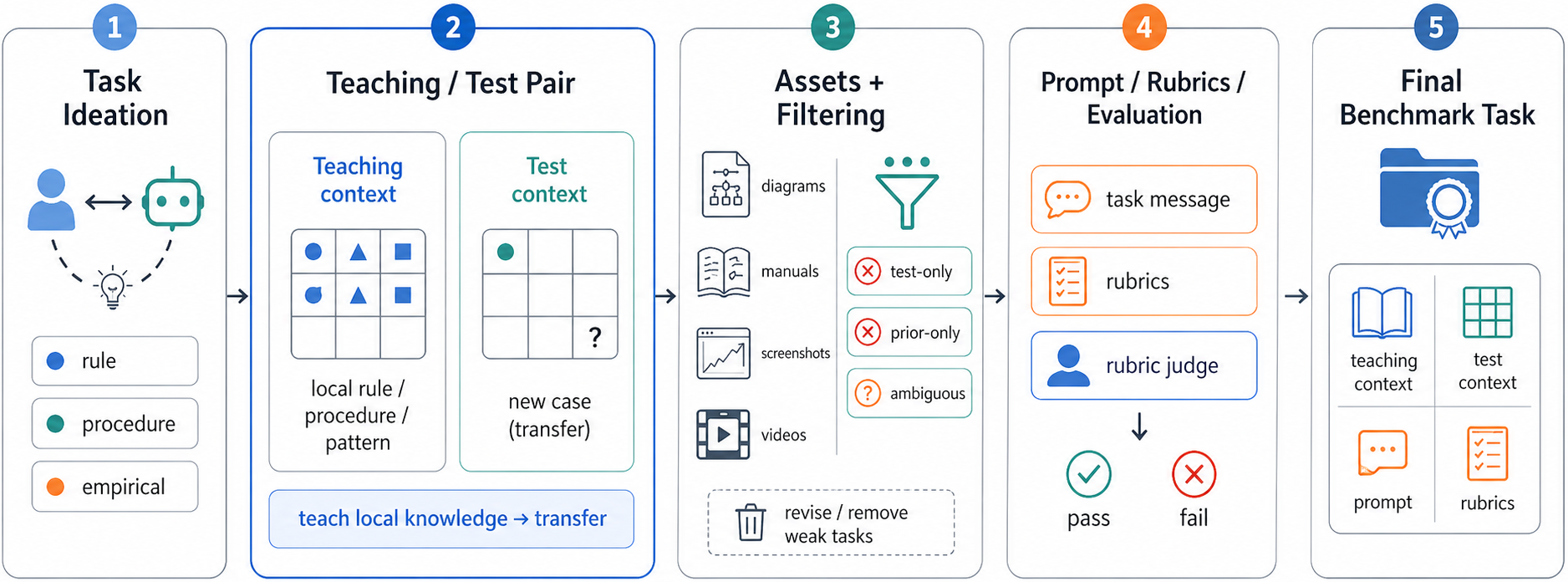}
    \caption{Construction pipeline for \benchmark{}. Candidate tasks are generated through a human-agent workflow, revised with leakage checks, and evaluated with rubrics designed to distinguish partial understanding from complete task success.}
    \label{fig:pipeline}
\end{figure}

As shown in Fig.~\ref{fig:pipeline}, \benchmark{} is constructed through a taxonomy-guided human--agent workflow. For each task, we first specify the target form of context learning---rule application, procedural execution, or empirical induction---and then instantiate it as a multimodal teaching--test pair. The central requirement is that the teaching context must contain the task-local source of truth, while the test context must require transferring that knowledge to a new case. This design principle is independent of surface domain: the same context-learning structure can be realized through diagrams, manuals, screenshots, frame sequences, videos, or synthetic visual states.


Task candidates are first proposed by both human designers and agents, forming a diverse candidate pool of potential MMCL tasks. Human curators then filter and refine this pool, selecting only candidates that are benchmark-worthy: the teaching context must contain a meaningful task-local source of truth, the test case must require transfer rather than direct matching, and the resulting task should reveal a nontrivial multimodal context-learning challenge.

For synthetic tasks, we use scripts or agent-guided generation to control the underlying rule structure and visual conditions. In addition to fully synthetic constructions, we deliberately create modified variants of existing content by instructing agents to alter selected rules, definitions, or specifications. These controlled modifications help ensure that the correct solution depends on the provided teaching context rather than on familiar priors.
For procedural or real-world tasks, we curate or search multimodal materials that define a local rule, procedure, or inspection criterion. In both settings, the goal is the same: the test instance should not be solvable by isolated visual recognition or direct example matching, but only by learning from the provided teaching context.

Filtering is an explicit part of the construction loop, especially for synthetic cases. We drop or revise candidates when the test instance appears solvable without the teaching context, when the intended rule can be recovered too easily from familiar priors, or when the visual evidence is too ambiguous to support reliable evaluation.
In practice, this filtering often requires regenerating or jointly revising the teaching context, the local rule or procedure it instantiates, and the corresponding test case so that the full task construction supports genuine context learning. The resulting benchmark therefore emphasizes low-contamination tasks in which success depends on the supplied multimodal context rather than on familiar priors or direct visual matching.

Messages and rubrics are written  after the multimodal assets are fixed. The message specifies the required output without revealing the hidden rule, and the rubrics decompose correctness into independently checkable criteria. Both are drafted with agent assistance and then manually reviewed to reduce leakage, ambiguity, and missing edge cases.

\subsection{Evaluation}
\label{sec:evaluation}
Following \citep{dou2026clbench}, each task is evaluated with strict binary scoring: a response receives a score of 1 only if it satisfies all rubrics associated with the task, and 0 otherwise. This all-or-nothing criterion ensures that partial understanding does not count as task success, particularly in cases where errors arise from incorrect localization, rule application, or state updates.

To complement binary task accuracy, we also report rubric-level accuracy, defined as the fraction of satisfied rubrics across all tasks. This provides a finer-grained view of model behavior, distinguishing near-miss cases from complete failures.

Evaluation is performed using an LLM-as-Judge protocol. The judge receives only the rubric file and the model’s textual response, without access to the original multimodal context. This isolates evaluation from additional perception errors and ensures that scoring depends solely on whether the response satisfies the specified criteria. \textcolor{black}{Across GPT-5.4, Gemini 3.1 Pro preview, and Claude Opus 4.6 used as rubric-only judges, pairwise rubric agreement remains above 90\%, suggesting that the scoring signal is reasonably stable across judge models; detailed results are reported in the appendix.}


%% file: sections/4_results.tex
\section{Results}
\label{sec:experiments}

\subsection{Main Model Evaluation}
\label{sec:main_eval}

We evaluate frontier multimodal language models on the 102-task \benchmark{}. The main comparison includes GPT-5.4 and GPT-5.4 thinking, Gemini 3.1 Pro preview, Kimi-K2.5 Thinking, Qwen3-6-Plus Thinking, and Claude Opus 4.6 thinking. Given the challenging nature of \benchmark{}, which requires models to read multimodal teaching context, infer task-local rules or procedures, and apply them to new visual test instances, we focus on strong multimodal models with extended reasoning. We report both strict task pass rate and rubric-level accuracy. Unless otherwise specified, models use their recommended or default decoding settings.

\input{tables/1_main_results}

GPT-5.4 thinking achieves the strongest overall result, solving 26.5\% of tasks. This is an 11.8-point improvement over GPT-5.4, but still leaves most tasks unsolved. Qwen3-6-Plus Thinking reaches 20.6\%, while Claude Opus 4.6 thinking reaches 19.6\%. Category results show different capability profiles: GPT-5.4 thinking is strongest on rule-system and empirical tasks, while Claude Opus 4.6 thinking is strongest on procedural tasks. Rubric-level accuracy is substantially higher than strict task success, indicating that many failures are partial successes in which a missed coordinate, row, manual step, or state update prevents full task completion.

\subsection{Diagnostic Oracle Ablation}
\label{sec:oracle_ablation}

To isolate where the context-to-answer pipeline fails, we run a diagnostic ablation on a visually grounded subset. This ablation is not part of the leaderboard. Instead, it compares five controlled input conditions: the original multimodal task, a no-teaching setting, a visual-state oracle that textualizes the relevant test-state details without directly giving the answer, a full-text oracle that adds textualized teaching evidence and test state, and a pure-text variant that removes the image inputs and expresses the task entirely in text. 
\textcolor{black}{We do not apply this conversion to the full benchmark, because many \benchmark{} tasks are intrinsically multimodal and do not admit a natural or lossless text-only reformulation. We therefore restrict the oracle study to a controlled subset of convertible tasks, mainly synthetic cases whose operative teaching and test evidence can be faithfully serialized into text without changing the underlying task semantics. The text-oracle conditions should accordingly be interpreted as diagnostic upper bounds, not as evidence that MMCL tasks can in general be reduced to text. Even within this controlled subset, the full-text and pure-text variants are not produced by directly transcribing the teaching context alone. They are constructed through an additional curation step that consolidates the task materials into a cleaner textual representation of the operative teaching and test evidence. Their gains therefore show how much performance improves when visual state and local context are made explicit in text, rather than implying that multimodal context is unnecessary or trivially replaceable.}

\input{tables/2_oracle_results}

As shown in Tab.~\ref{tab:oracle_ablation}, the no-teaching setting reaches 0.0\% overall, supporting our construction goal that the selected tasks require context rather than only prior knowledge. The visual-state oracle improves overall performance from 37.5\% to 62.5\%, indicating that visual-state extraction is a measurable bottleneck: in several cases, correctly recovering coordinates, colors, or object locations is enough to make the downstream local rule application succeed. Full-text and pure-text variants both reach 75.0\% overall, suggesting that many failures originate in visual reading or multimodal localization rather than in the abstract context-learning step itself. At the same time, the gap between the text-oracle conditions and perfect performance shows that perception is not the only limitation. Some remaining failures are dominated by long-chain state simulation and rule execution, while a small number of near misses are better interpreted as strict wording or rubric-coverage issues rather than substantive reasoning errors.

%% file: tables/1_main_results.tex
\begin{table}[t]
    \centering
    \small
    \caption{Main results on \benchmark{}. ``Overall'' reports strict task pass rate under all-rubric scoring, and ``Rubrics'' reports rubric-level accuracy. Category columns report strict task pass rate within each top-level category.}
    \label{tab:main_results}
    \begin{tabularx}{\linewidth}{Xrrrrr}
        \toprule
        \multirow{2}{*}{Model} & \multirow{2}{*}{Overall} & \multirow{2}{*}{Rubrics} & \multicolumn{3}{c}{Task Categories} \\
        \cmidrule(lr){4-6}
        & & & Rule & Procedural & Empirical \\
        \midrule
        GPT-5.4 & 14.7\% & 59.9\% & 22.4\% & 9.4\% & 4.8\% \\
        Gemini 3.1 Pro preview & 15.7\% & 67.6\% & 16.3\% & 12.5\% & 19.0\% \\
        Kimi-K2.5 Thinking & 16.7\% & 64.1\% & 20.4\% & 21.9\% & 0.0\% \\
        Claude Opus 4.6 thinking & 19.6\% & 66.0\% & 20.4\% & \textbf{25.0\%} & 9.5\% \\
        Qwen3-6-Plus Thinking & 20.6\% & 67.6\% & 28.6\% & 18.8\% & 4.8\% \\
        GPT-5.4 thinking & \textbf{26.5\%} & \textbf{76.6\%} & \textbf{32.7\%} & 18.8\% & \textbf{23.8\%} \\
        \bottomrule
    \end{tabularx}
\end{table}

    

%% file: tables/2_oracle_results.tex
\begin{table}[t]
    \centering
    \small
\caption{Diagnostic oracle ablation on a visually grounded subset using GPT-5.4 thinking. Teach-V and Test-V denote the original visual or multimodal teaching context and test instance; Teach-T and Test-T denote textualized oracle versions of the corresponding evidence. Overall uses the strict all-rubric task criterion; Rubrics reports rubric-level accuracy.}
    \setlength{\tabcolsep}{4pt}
    \begin{tabular}{lcccccc}
        \toprule
        \multirow{2}{*}{Condition} & \multicolumn{4}{c}{Input components} & \multirow{2}{*}{Overall} & \multirow{2}{*}{Rubrics} \\
        \cmidrule(lr){2-5}
        & Teach-V & Test-V & Teach-T & Test-T & & \\
        \midrule
        Original & $\checkmark$ & $\checkmark$ & -- & -- & 37.5\% & 79.3\% \\
        No teaching & -- & $\checkmark$ & -- & -- & 0.0\% & 41.4\% \\
        Visual-state oracle & $\checkmark$ & $\checkmark$ & -- & $\checkmark$ & 62.5\% & 85.0\% \\
        Full-text oracle & $\checkmark$ & $\checkmark$ & $\checkmark$ & $\checkmark$ & 75.0\% & 89.3\% \\
        Pure-text & -- & -- & $\checkmark$ & $\checkmark$ & 75.0\% & 87.1\% \\
        \bottomrule
    \end{tabular}
    \label{tab:oracle_ablation}
\end{table}

%% file: sections/5_analysis.tex
\section{Analysis}
\label{sec:analysis}

\subsection{Why Multimodal Context Learning is Difficult}
\label{sec:why_mmcl_hard}

We view multimodal context learning as a context-to-answer pipeline. 
Our current results suggest that failures can enter at each stage. The diagnostic oracle ablation in Table~\ref{tab:oracle_ablation} also confirms that the benchmarked tasks are genuinely context-dependent: removing the teaching context reduces overall performance from 37.5\% to 0.0\%.

\begin{figure}[t]
    \centering
    \includegraphics[width=\linewidth]{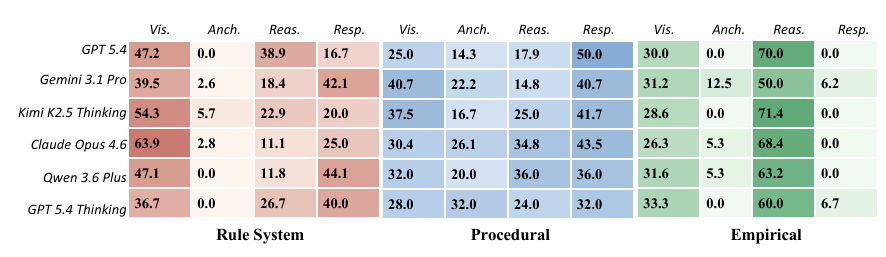}
    \caption{Category-conditioned 4-way error-pattern heatmap over failed responses. Vis, Anch, Reas, and Resp denote visual evidence extraction, context anchoring, context reasoning, and response construction, respectively.}
    \vspace{-15pt}
    \label{fig:error_pattern_heatmap}
\end{figure}

Rather than hand-designing a fine-grained taxonomy in advance, we use a model-assisted bottom-up coding pass over many failed responses to recover the dominant recurring error types. This process stabilizes around four coarse labels: context anchoring, visual evidence extraction,  context reasoning, and response construction. Figure~\ref{fig:error_pattern_heatmap} reports these emergent labels by top-level task category, because the benchmark reveals a strong association between task family and dominant error type.

\vspace{2pt}
\noindent \textbf{Context anchoring failures: finding the wrong place.}
Even when the relevant information is present and visible, the model may fail to locate the right part of the context. This is especially severe for long procedural contexts. Assembly instructions are the hardest subcategory in \benchmark{}, with only 4.2\% average pass across the eight complete settings. These tasks require aligning a current physical state with the correct manual page, step, part, and visual evidence. This is related to long-context retrieval, but it is harder than text retrieval alone: visual pages, diagrams, part photos, and the test image must be matched under viewpoint changes and partial assembly states. In the heatmap, context anchoring rises substantially in Procedural Task Execution relative to the other two main categories.

\begin{figure}[t]
    \centering
    \includegraphics[width=\linewidth]{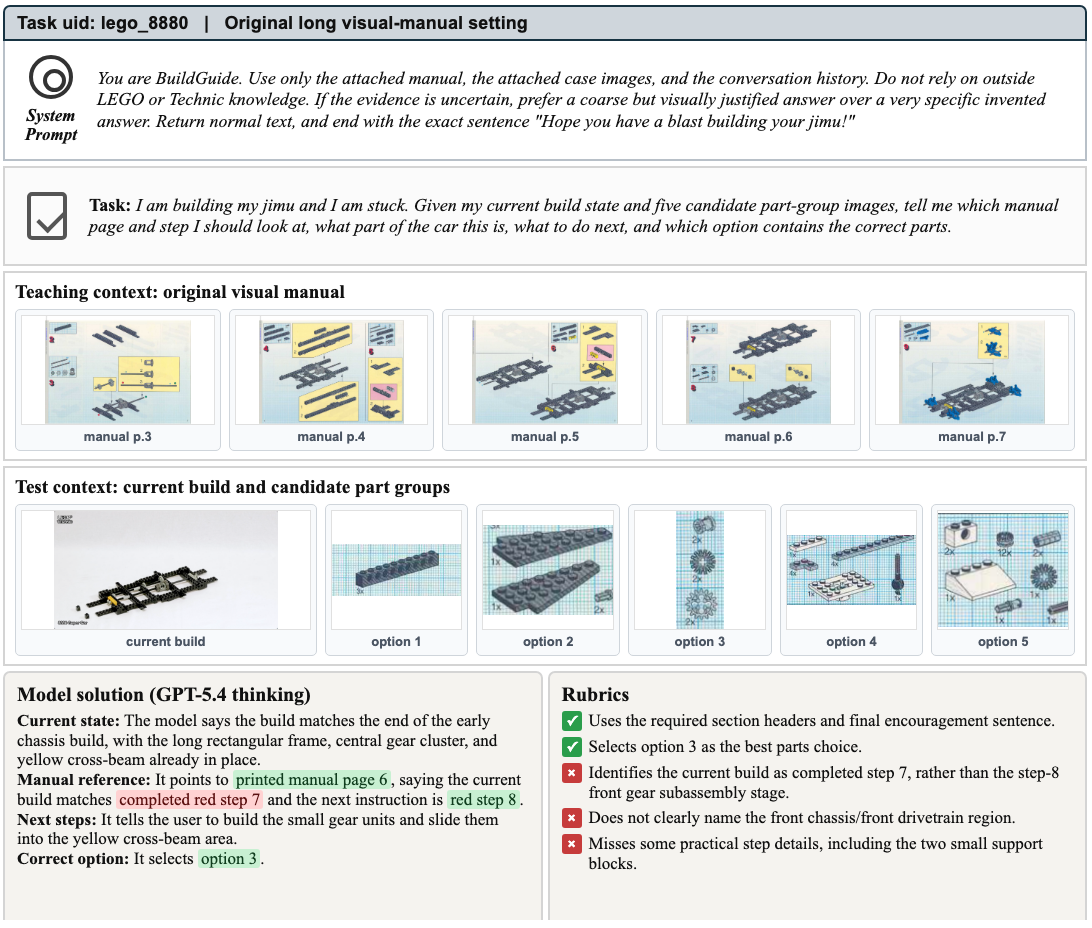}
    \caption{Example of a visual-manual localization failure. GPT-5.4 thinking fails because it misaligns the current build state with the exact step.}
    \vspace{-15pt}
    \label{fig:case_lego_8880}
\end{figure}

\input{tables/3_crop_ablation}

Figure~\ref{fig:case_lego_8880} shows a concrete original-setting example: the model locates the broad early-chassis region and even chooses the correct candidate part group, but the answer still fails because it blurs the boundary between the completed previous step and the current front gear-subassembly step. Table~\ref{tab:manual_localization_ablation} supports the same interpretation at the subset level. A pointer to the relevant page or step raises Rubrics from 59.0\% to 70.5\%, indicating that many local requirements become easier once the model is told where to look. However, strict Overall remains 0.0\%, so localization is not the only bottleneck. Cropping can produce a strict success, raising Overall to 12.5\%, but it is less stable at the rubric level than pointer guidance, suggesting that cropped views may remove useful procedural context for source fitting, next-step judgment, or part-state interpretation. 


\vspace{2pt}
\noindent \textbf{Visual perception limitations: seeing poorly.}
Some failures occur before reasoning begins: the model does not extract the necessary visual detail from the input. This can happen when objects are small, layouts are dense, colors are subtle, or images are low-resolution or visually noisy. Large images, screenshots, videos, and manual pages may also be resized, compressed, or tokenized before the model reasons over them, so fine details can be lost in the visual representation. In the oracle ablation, adding a textualized visual-state oracle improves overall performance from 37.5\% to 62.5\% and Rubrics from 79.3\% to 85.0\%. The gains suggest that correctly reading coordinates, colors, object identities, or locations can be enough to make the downstream local rule application succeed. The heatmap is consistent with this interpretation: visual evidence extraction remains a frequent failure label in both rule-system and procedural tasks.

\vspace{2pt}
\noindent \textbf{Context reasoning failures: learning or applying the wrong rule.}
Some failures remain even after the model has perceived and localized the relevant context. This is not uniquely visual; it is the core context-learning bottleneck. In the 4-way heatmap, context reasoning is substantial across all three main categories and becomes the dominant label in the empirical tasks. In rule-system settings, this often means that the model finds the relevant evidence but still applies the taught rule incorrectly once it has been found. When the task-local rule conflicts with pretrained knowledge, models may default to a familiar prior rather than follow the provided context. 
Another reasoning failure is rule-type confusion: the model detects that a pattern exists but represents the wrong kind of rule, such as direction rather than distance, layer count, or shape.


The same label takes a stronger form in empirical discovery and induction. There, the main difficulty is not simply following an explicit rule, but inferring the correct latent rule from examples and then transferring it to a new case. This is the capability needed for inspection-style settings such as anomaly detection, industrial defect detection, and sequence-based visual reasoning. In the current benchmark, empirical discovery and induction is the hardest main category, and the GPT-5.4 high-reasoning ablation improves empirical pass rate from 4.8\% to 23.8\%, suggesting that extended comparison and hypothesis formation help but remain insufficient. Dynamic tasks are especially difficult because the answer depends on a sequence of intermediate states rather than a single final image. For example, in a gear-chain style task, the model must infer how motion should propagate through connected parts, stop propagation at a broken component, and update every downstream component accordingly. A response can look almost correct locally but still fail if one intermediate state is updated under the wrong learned rule. The oracle ablation reinforces this point: even under full-text and pure-text variants, some failures remain because the task requires multi-step state simulation, such as repeated selection, exclusion of previously used elements, spatial updates, or decay-like transitions. 

\vspace{2pt}
\noindent 
\textbf{Response construction failures: knowing but not fully delivering.}
A smaller but recurring share of failures occurs after the core reasoning is mostly in place. These responses omit one required exception, fail to make a key condition explicit, or do not match the requested output structure closely enough to satisfy the rubric. The heatmap shows that this label appears across categories, particularly in rule-system and procedural tasks. We therefore keep response construction as a separate coarse label, since some failures arise not from mislearning the context itself but from incomplete or insufficiently precise final outputs.


%% file: tables/3_crop_ablation.tex
\begin{wraptable}[13]{r}{0.42\columnwidth}
    \centering
    \small
    \caption{Manual localization ablation using GPT-5.4 thinking. 
    ``Pointer'' gives the relevant page number, and ``Cropped'' gives a cropped view of  relevant content. }
    \setlength{\tabcolsep}{4pt}
    \begin{tabular}{lcc}
        \toprule
        Condition & Overall & Rubrics \\
        \midrule
        Original & 0.0\% & 59.0\% \\
        Pointer  & 0.0\% & 70.5\% \\
        Cropped  & 12.5\% & 63.8\% \\
        \bottomrule
    \end{tabular}
    \label{tab:manual_localization_ablation}
\end{wraptable}


%% file: sections/7_conclusion.tex
\section{Conclusion}
\label{sec:conclusion}

We introduced \benchmark{}, a 102-task benchmark for multimodal context learning from visual and mixed-modality teaching context. Across rule-system, procedural, and empirical discovery tasks, current frontier models remain far from saturation: GPT-5.4 thinking achieves only 26.5\% strict task success. Our analyses show that this gap cannot be explained by a single bottleneck. Success depends on using the teaching context itself, while failures arise from a combination of context anchoring,
visual state extraction, and downstream rule or procedure application. These results motivate multimodal systems with stronger grounding, intermediate-state verification, and long-context visual reasoning.

\textbf{Implications for multimodal systems.}
These findings suggest that progress in multimodal context learning will require more than stronger end-to-end reasoning alone. Future systems may also need better visual state extraction, long-context grounding, and intermediate-state verification. \benchmark{} therefore points to a broader systems challenge rather than a bottleneck in final-answer reasoning alone.

\textbf{Limitations and future scale.}
The current benchmark remains limited in scale, and the diagnostic oracle and manual-localization studies cover only small subsets designed to isolate mechanisms rather than serve as separate leaderboards. We do not yet report human or expert baselines, which would further calibrate task solvability and difficulty. In addition, evaluation relies on a rubric-only LLM judge: removing multimodal judge confounds improves consistency, but rubric wording can still affect borderline cases. A useful next step is to expand task coverage while preserving both controlled tasks for mechanism-level diagnosis and grounded practical tasks that reflect real deployment settings.




%% file: sections/8_appendix.tex
\newpage
\appendix

\section{Supplementary Evaluation Details}
\label{app:eval_details}

\subsection{API and Agent Comparisons}
\label{app:agent_comparison}

Table~\ref{tab:api_agent_comparison} reports matched direct-model and agentic settings for the two base models where both were run. We keep these comparisons in the appendix because agentic execution changes computation, retrieval behavior, and visual rechecking, making it a different evaluation regime from the main direct-model results.

\begin{table}[h]
    \centering
    \small
    \caption{Matched API and agent settings on the benchmark. }
    \label{tab:api_agent_comparison}
    \begin{tabular}{llrrrrr}
        \toprule
        Model & Setting & Overall & Rubrics & Rule & Procedural & Empirical \\
        \midrule
        Claude Opus 4.6 thinking & API & 19.6\% & 66.0\% & 20.4\% & 25.0\% & 9.5\% \\
        Claude Opus 4.6 thinking & Agent & 18.6\% & 66.2\% & 18.4\% & 25.0\% & 9.5\% \\
        GPT-5.4 thinking & API & 26.5\% & 76.6\% & 32.7\% & 18.8\% & 23.8\% \\
        GPT-5.4 thinking & Agent & 31.4\% & 80.3\% & 38.8\% & 21.9\% & 28.6\% \\
        \bottomrule
    \end{tabular}
\end{table}

The two matched comparisons behave differently. For Claude Opus 4.6 thinking, the agent setting is essentially flat relative to the direct-model setting, with a slight decrease in Overall and nearly identical rubric-level accuracy. For GPT-5.4 thinking, the agent setting improves both Overall and Rubrics, suggesting that rechecking and longer interaction are more beneficial for this model. Even so, the absolute results remain far from saturation, so agentic scaffolding does not remove the core MMCL bottlenecks documented in the main text.

\subsection{Diagnostic Subset Definitions}
\label{app:diagnostic_subsets}

\paragraph{Oracle ablation subset.}
The oracle ablation in the main paper is run on a visually grounded subset of $n{=}24$ tasks. We restrict this subset to tasks whose operative teaching and test evidence can be faithfully textualized without changing the underlying task semantics. In practice, this means mainly controlled synthetic tasks spanning rule-system and empirical settings, rather than long-form procedural tasks whose core difficulty depends on extended visual navigation through manuals or workflow context.

\paragraph{Manual-localization subset.}
The manual-localization ablation in the main paper is run on an instruction-manual subset. These tasks come from the procedural category, specifically assembly-style cases in which the model must align a current physical state with the correct manual page, step, and local visual evidence.

\paragraph{Context-necessity filtering for synthetic tasks.}
For synthetic tasks, we explicitly test whether the teaching context is necessary for success. We remove the teaching context, restart evaluation with a fresh agent that sees only the test case and prompt, and check whether the task can still be solved directly. If the no-context run succeeds, we revise the task. If a task remains solvable without teaching context after up to three rounds of revision, including manual intervention, we drop it from the benchmark. In practice, these dropped cases are usually either too easy or too difficult to construct in a way that cleanly isolates multimodal context learning.

\subsection{Alternative Judge Models}
\label{app:alternative_judges}

\paragraph{Judge consistency across rubric-only judges.}
Across GPT-5.4, Gemini 3.1 Pro preview, and Claude Opus 4.6, pairwise rubric agreement remains above 91\%, while strict task-level agreement is slightly lower because a single disputed rubric can flip the final pass/fail decision. In other words, changing the judge shifts absolute pass rates, but much of the rubric-level scoring signal remains shared.

\begin{table}[h]
    \centering
    \small
    \caption{Pairwise agreement among the three rubric-only judges.}
    \label{tab:judge_consistency_summary}
    \begin{tabular}{lccc}
        \toprule
        Metric & GPT vs Gemini & GPT vs Claude & Gemini vs Claude \\
        \midrule
        Task agreement & 90.5\% & 87.1\% & 90.7\% \\
        Rubric agreement & 92.5\% & 91.2\% & 92.5\% \\
        \bottomrule
    \end{tabular}
\end{table}

\subsection{Some Factors Are Less Decisive on Their Own}
Not every apparent complexity source reliably predicts failure. Rule complexity alone is often insufficient to explain difficulty when the visual state is clean and easy to localize, while even simple rules can remain challenging when the evidence is dense, dispersed, or hard to anchor. Output length is likewise not a sufficient explanation: tasks with detailed responses may still be easy when the relevant evidence is explicit, whereas short-answer tasks can remain difficult when they require precise retrieval, grounding, and state tracking. Finally, strict rubric wording may affect isolated near misses, but it does not explain the overall performance gap.

\subsection{Broader impacts.}
\benchmark{} may have positive impact as a diagnostic benchmark for multimodal systems that must learn from local visual or mixed-modality context, including assistive, educational, and workflow-support settings. By identifying failures in visual evidence extraction, context anchoring, and downstream reasoning, it may also help improve the reliability of multimodal systems in practical tasks.

At the same time, better multimodal context learning could be misused in surveillance, deceptive automation, or highly capable agents that interpret visual instructions or operational materials more effectively than intended. In addition, benchmarks built from practical materials may inherit licensing, privacy, or deployment-related risks if released without sufficient review. We therefore view careful asset curation, licensing checks, and responsible release practices as important parts of future benchmark expansion.

\subsection{Licenses for existing assets.}
The paper uses existing models, benchmark references, and source materials only when their usage is permitted under the corresponding public licenses or terms of use. We credit the original creators of these assets in the paper and supplementary material, and we do not intentionally include third-party assets whose licensing status is unclear. For benchmark construction, we restrict reused or adapted source materials to assets with open or otherwise usable licenses, and future public release will include a consolidated asset-by-asset license summary.

\subsection{Execution time and concurrency.}
A complete benchmark run typically requires roughly two to three hours of continuous API execution for one model, depending on model latency and temporary rate limits. Our evaluation pipeline uses up to eight concurrent worker processes to keep throughput stable across the 102 tasks. We report this information to clarify the practical cost of reproducing the benchmark evaluations, even though the paper does not involve model training.


\section{Error-Pattern Annotation Details}
\label{app:error_pattern_plan}

Our error analysis uses the same four coarse mechanisms reported in the main text: visual evidence extraction, context anchoring, context reasoning, and response construction. In the classifier prompt below, these are represented as the JSON labels \texttt{visual\_evidence\_extraction}, \texttt{context\_anchoring}, \texttt{context\_reasoning}, and \texttt{response\_construction}.

\subsection{Annotation Workflow}

Figure~4 in the main paper is computed over the failed responses from the six main direct-model runs. We first carried out an agent-based bottom-up coding pass over these failed responses to identify the dominant recurring mechanisms. After this exploratory pass stabilized, we fixed the four-label scheme used in the main paper and applied the classifier prompt below to assign labels to individual failures. Each failed response is allowed to receive up to three labels, since a single error can reflect multiple interacting breakdowns in the context-to-answer pipeline. We then manually reviewed a subset of the agent-assigned labels to check that the boundary rules were behaving as intended and refined the instructions when recurring ambiguities appeared.

\subsection{Boundary Rules}

The most important distinctions are operational rather than semantic. The boundary between \texttt{visual\_evidence\_extraction} and \texttt{context\_anchoring} is determined by the \emph{locus of failure}: when the model reaches the correct evidence source but misreads local visual content, we assign \texttt{visual\_evidence\_extraction}; when it fails to retrieve or align the correct page, frame, region, or example in the first place, we assign \texttt{context\_anchoring}. The boundary between \texttt{context\_anchoring} and \texttt{context\_reasoning} is determined by whether the relevant evidence was found: if the model reaches the right evidence source but still applies the rule, procedure, or inferred pattern incorrectly, we assign \texttt{context\_reasoning}. This label therefore covers both explicit rule misuse and failures to induce the correct latent rule from examples or temporal evidence. 
Finally, \texttt{response\_construction} is used when the model appears to have substantially recovered the relevant rule or judgment but fails to fully realize it in the final answer. This category therefore includes not only formatting mismatches, but also constraint-adherence errors such as omitting a required exception, leaving out a key condition, or failing to satisfy the requested output structure precisely enough to meet the rubric.


\subsection{Classifier Prompt}

The fixed classifier prompt is shown below. In practice, task-specific fields are filled automatically from each task directory and paired judge file.

\begin{Verbatim}[fontsize=\scriptsize,frame=single]
You are labeling one failed model response for task-level error patterns in
MM-CL-Bench.

Use only these task files:
- messages.json
- response/<model>.txt
- rubrics.json
- judge/<model>_judge_by_*.json

Do not use images or any other files.

Allowed labels only:
- visual_evidence_extraction
- context_anchoring
- context_reasoning
- response_construction

No primary label.

Label guidance:
- visual_evidence_extraction: the response misreads visually grounded details
  such as piece type, count, row/column, color, OCR content, or demonstrated
  local state after attending to the right evidence source.
- context_anchoring: the response fails to find the relevant page, frame,
  region, row, column, or other needed evidence in the provided task context.
- context_reasoning: the response reaches the relevant evidence but uses the
  learned rule, procedure, or inferred pattern incorrectly. This includes
  prior override, incomplete rule application, and failures to induce the
  correct latent rule from examples or temporal evidence.
- response_construction: the response largely reflects the right reasoning but
  fails to satisfy the rubric because it omits a required exception, leaves a
  key condition implicit, or does not match the requested output format.

Instructions:
- Assign 1 to 3 labels.
- Use context_reasoning when the relevant evidence has been found but the
  learned rule, procedure, or inferred pattern is still applied incorrectly.
- If the response mainly fails because it did not find the right page, frame,
  region, or board location, prefer context_anchoring.
- Use response_construction when the core judgment appears substantially
  correct but the final answer remains incomplete or insufficiently precise.
- Keep the reason short and grounded in judge evidence.
- Return JSON only.
\end{Verbatim}

\clearpage
\section{Additional Case Examples}
\label{app:case_examples}

\begin{figure}[h]
    \centering
    \includegraphics[width=\linewidth]{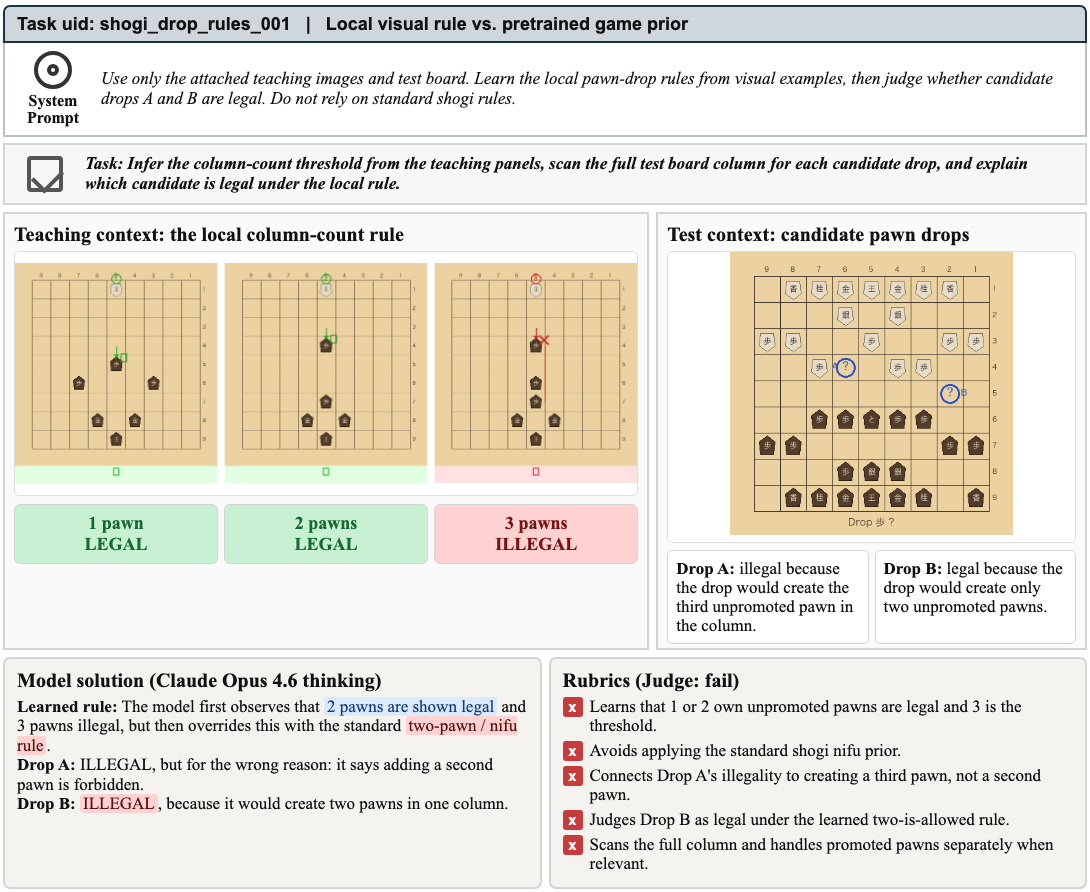}
    \caption{Example of a context-reasoning failure caused by prior override. The teaching context defines a local shogi-like pawn-drop rule: one or two own unpromoted pawns in a column are legal, while three is illegal. Claude Opus 4.6 thinking instead applies the standard two-pawn prior and judges Drop B as illegal, even though it should be legal under the task-local rule.}
    \label{fig:case_shogi_drop_rules}
\end{figure}

\begin{figure}[h]
    \centering
    \includegraphics[width=\linewidth]{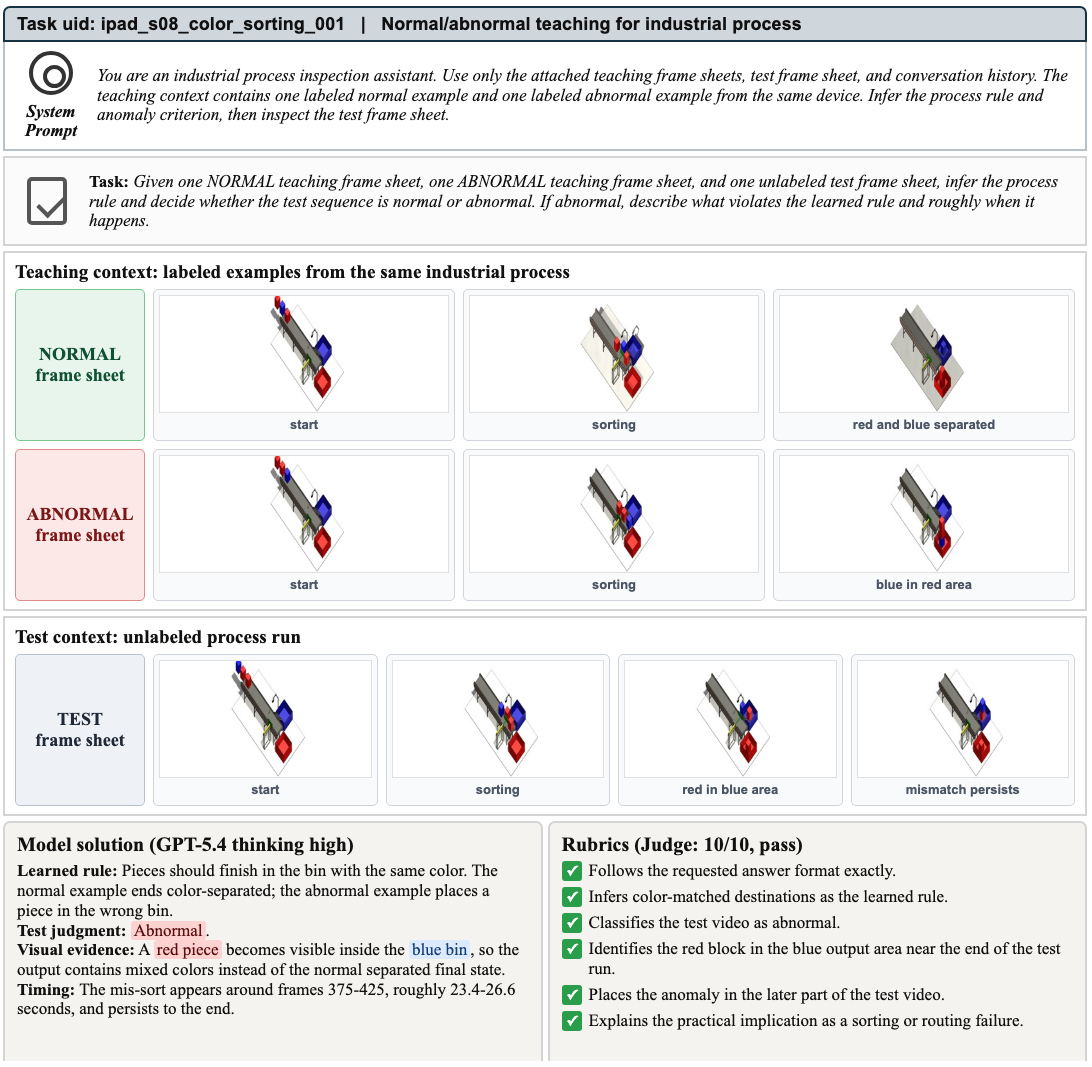}
    \caption{Industrial inspection example from the empirical discovery and induction category. The task provides one normal teaching frame sheet and one abnormal teaching frame sheet from the same process, then asks the model to classify a new test sequence. GPT-5.4 thinking correctly infers the local color-sorting criterion, identifies the red piece entering the blue output area, and passes all rubrics.}
    \label{fig:case_ipad_s08}
\end{figure}

\begin{figure}[h]
    \centering
    \includegraphics[width=\linewidth]{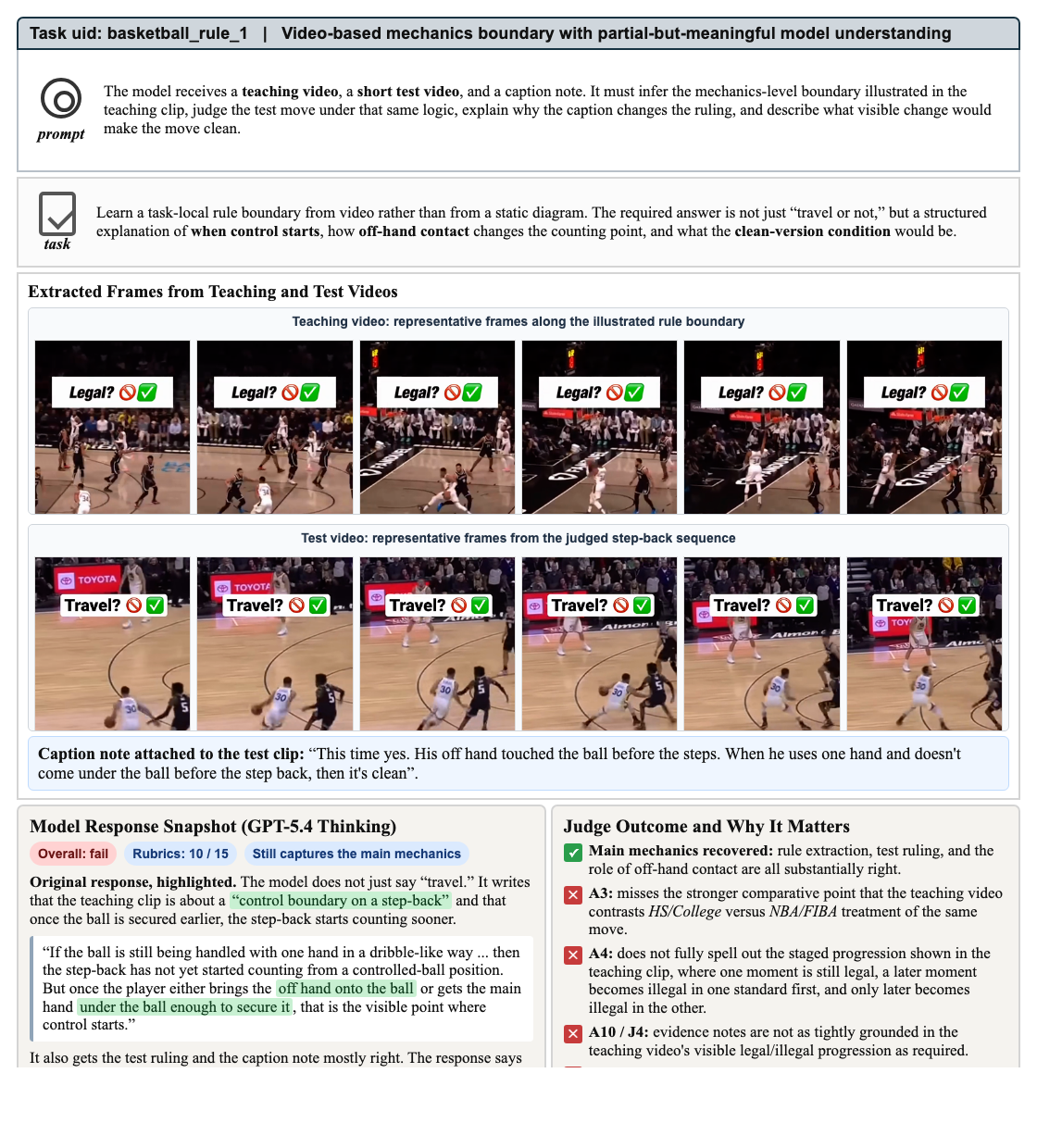}
    \caption{Example of a video-based context-learning failure with substantial partial understanding. The model recovers the core control-boundary idea and the role of off-hand contact, but misses the finer staged comparison encoded in the teaching video, showing that the benchmark distinguishes coarse intuition from fully grounded use of task-local visual evidence.}
    \label{fig:case_basketball}
\end{figure}